\documentclass[conference]{IEEEtran}
\IEEEoverridecommandlockouts
\usepackage{amsmath,amssymb,amsfonts}
\usepackage{algorithmic}
\usepackage{graphicx}
\usepackage{textcomp}
\usepackage{xcolor}
\def\BibTeX{{\rm B\kern-.05em{\sc i\kern-.025em b}\kern-.08em
 T\kern-.1667em\lower.7ex\hbox{E}\kern-.125emX}}

\usepackage{multirow}
\usepackage{subcaption}
\usepackage{url}
\usepackage{soul}
\usepackage{hyperref}
\usepackage{footnote}
\usepackage{tablefootnote}
\usepackage{biblatex}

\usepackage{graphicx}
\usepackage{subcaption}
\usepackage{makecell}

\PassOptionsToPackage{hyphens}{url}\usepackage{hyperref} 

\usepackage{csquotes}
\title{Can pre-trained Transformers be used in detecting complex sensitive sentences? - A Monsanto case study}

\author{

\IEEEauthorblockN{Roelien C. Timmer}
\IEEEauthorblockA{\textit{UNSW Sydney, Data61} \\
\textit{Cyber Security CRC}\\Sydney, Australia \\r.timmer@unsw.edu.au}\and

\IEEEauthorblockN{David Liebowitz}
\IEEEauthorblockA{\textit{Penten} \\\textit{UNSW Sydney}\\Canberra, Australia\\david.liebowitz@penten.com}\and

\IEEEauthorblockN{Surya Nepal}
\IEEEauthorblockA{\textit{Data61} \\\textit{Cyber Security CRC}\\Sydney, Australia \\surya.nepal@data61.csiro.au}\and

\IEEEauthorblockN{Salil S. Kanhere}\IEEEauthorblockA{\textit{UNSW Sydney} \\\textit{Cyber Security CRC}\\Sydney, Australia \\salil.kanhere@unsw.edu.au}

}

\begin{document}
\maketitle

\begin{abstract}
Each and every organisation releases information in a variety of forms ranging from annual reports to legal proceedings. Such documents may contain sensitive information and releasing them openly may lead to the leakage of confidential information. Detection of sentences that contain sensitive information in documents can help organisations prevent the leakage of valuable confidential information. This is especially challenging when such sentences contain a substantial amount of information or are paraphrased versions of known sensitive content. Current approaches to sensitive information detection in such complex settings are based on keyword-based approaches or standard machine learning models. In this paper, we wish to explore whether pre-trained transformer models are well suited to detect complex sensitive information. Pre-trained transformers are typically trained on an enormous amount of text and therefore readily learn grammar, structure and other linguistic features, making them particularly attractive for this task. Through our experiments on the Monsanto trial data set, we observe that the fine-tuned Bidirectional Encoder Representations from Transformers (BERT) transformer model performs better than traditional models. We experimented with four different categories of documents in the Monsanto dataset and observed that BERT achieves better F2 scores by 24.13\% to 65.79\% for GHOST, 30.14\% to 54.88\% for TOXIC, 39.22\% for CHEMI, 53.57\% for REGUL compared to existing sensitive information detection models.
\end{abstract}

\begin{IEEEkeywords}
sensitive information detection, NLP, transformers, BERT, Monsanto Trial
\end{IEEEkeywords}

\section{Introduction}\label{introduction}
Sensitive information detection is more important than ever as many organisations store confidential documents digitally. Having a sensitive information detection tool in place helps users avoid accidentally leaking valuable information.

Over the past few years, a lot of research has been conducted on the detection of sensitive information using machine learning (ML) models. For example, identifying personal information such as name and date of birth in medical records \cite{pablos2020sensitive, perez2019vicomtech} and classifying documents based on their sensitivity such as labelling military documents sensitive and corporate documents non-sensitive \cite{xu2019detecting, lin2020sensitive}. These works are focused on unstructured data but do not address how complex information should be handled. An example of a complex sensitive sentence from the legal proceedings against the agricultural chemical company Monsanto from 2017 \cite{mchenry2018monsanto}, which we use in this work is: 
\begin{displayquote}
``Please delete it and all attachments from any servers, hard drives or any other media."
\end{displayquote}
This sentence was labelled as sensitive by the lawyers working on the legal case as they identified that the sentence relates to ghostwriting, i.e., to write for and in the name of another, undertaken by Monsanto employees. Since it does not explicitly contain the word {\it ghostwriting}, traditional methods cannot detect this sentence as sensitive.

The work by Neerbeck et al. in \cite{neerbek2020real} is the only work so far that has focused on how the context of a sentence contributes to the decision of whether that sentence contains sensitive information. In this work, the Monsanto Trial documents are labelled and analysed at the granularity of individual sentences, also referred to as the {\it golden} data set. See Section \ref{data_set} for more information about silver and golden labels. Each sentence is manually labelled as either sensitive or non-sensitive. The authors also tested four different models on the golden Monsanto data set sentences to detect sensitive information. The results presented in the paper raises a question: can we use a pre-trained transformer for such a task?

In this work, we answer the question by exploring and evaluating whether a pre-trained BERT transformer \cite{devlin2018bert} can be used to detect complex sensitive information. The motivation for the use of BERT comes from the following three facts: 

\begin{enumerate}
 \item Many researchers have shown that the BERT pre-trained transformers achieve good performance in many NLP tasks such as General Language Understanding Evaluation (GLUE) benchmark \cite{wang2018glue}, Stanford Question Answering data set (SQuAD) \cite{rajpurkar2016squad} and Situations With Adversarial Generations (SWAG) \cite{zellers2018swag} data set.
 \item Transformers allow for parallelisation and are therefore faster \cite{devlin2018bert}.
 \item Using a transfer learning approach with a pre-trained language model like BERT requires substantially fewer data to obtain the same quality of results compared to training a model from scratch because we only need to fine-tune the BERT model. Having a model that can handle small data sets is important because, like in the Monsanto case, organisations often have small labelled data sets.
\end{enumerate}

Our evaluation shows that by using BERT, high accuracy, precision, recall and F1 scores can be achieved even though the Monsanto data set is small. 

In short, our contributions are:
\begin{enumerate}
 \item We show that with pre-trained transformers we can achieve Accuracy, Precision, Recall, F1 and F2 scores for four different golden data sets, GHOST, TOXIC, CHEMI and REGUL, that are better than existing state-of-the-art techniques. Detailed information about this data set can be found in Section \ref{data_set}. 
 \item We show that the classification results of fine-tuning BERT on a sensitive information corpus are highly dependent on the initial values used in training. This means that whenever BERT is fine-tuned on a relatively small corpus, care should be taken to estimate a confidence interval.
\end{enumerate}

The paper is organised as follows. In Section \ref{related_work} we discuss existing sensitive information detection methods, such as the Inference Rule, C-sanitized and methods based on machine learning models. We also provide an overview of transformers in general and specifically BERT, which is used in our sensitive information detection method. In Section \ref{proposed_model} we describe our sensitive information detection method based on BERT. In Section \ref{evaluation} we discuss the evaluation results with, in Subsection \ref{data_set} the Monsanto data set we use, in Subsection \ref{metrics} the metrics we use for evaluation, in Subsection \ref{results} the results and in Subsection \ref{sec:LSTM} a discussion. We conclude this work in Section \ref{sec:conclusion} with a conclusion.

\section{Related Work}\label{related_work}

Automatic sensitive information detection used to be done with keyword-based models and more recently ML models have been utilised. In this work, we test whether a new type of ML model, the pre-trained Transformer, can be used to detect sensitive information. In Section \ref{inference_rule} and Section \ref{c-sanitized}, we discuss two keyword-based models, the Inference Rule and C-sanitized. In Section \ref{sec:LSTM} and Section \ref{sec:recnn}, we discuss how the LSTM and the RecNN ML models have been applied to detect sensitive information. In Section \ref{sec:transformers}, we discuss pre-trained transformers which we will test in this paper to detect sensitive information.

\subsubsection{Inference Rule}\label{inference_rule}

Chow \textit{et al.} \cite{cho2008detectingprivacy} introduced the Inference Rule to detect sensitive information in 2008. This approach uses corpus-based association rules, based on word co-occurrence. The Inference Rule is a sophisticated version of association rule mining \cite{agrawal1994fast}. It is based on the probability that sensitive information co-occurs with other words. The most simple case is that word $w_1$ implies confidential information $c$:
\begin{equation} \label{eq:comb_Inference Rule}
 w_1 \Rightarrow c 
\end{equation}
The inference rule can be more complex, combining conjunction and disjunction. For example, if a piece of text contains either $w_1$ or $w_2$ in combination with $w_3$, it implies confidential information $c$:

\begin{equation} 
 (w_1 \vee w_2) \wedge w_3 \Rightarrow c 
\end{equation}

On the training data set, the Inference Rule model learns the inferences between the sensitive information and other information. In this work, we use in line with \cite{neerbek2020real}, simple rules and a fixed confidence cutoff. The cutoff values will be based on the validation data set.

A disadvantage of the Inference Rule is that its knowledge is limited to the training corpus. Suppose that we have a rule as in Eq. \ref{eq:comb_Inference Rule}. Now imagine, that $w_1$ has a synonym $w4$ and therefore also implies sensitive information, $w_4 \Rightarrow c$. If $w4$ does not appear in the training corpus, the Inference Rule will not predict $w4$ to be sensitive as the Inference Rule does not make use of semantic similarity via word embeddings. The advantage of using a fine-tuned pre-trained Transformer model such as BERT is that it has been trained on a much larger corpus and therefore knows a great deal about a language, including how words are related.

\subsubsection{C-sanitized}\label{c-sanitized}
C-sanitized detection is based on point-wise mutual information (PMI) \cite{sanchez2016c}. It calculates the probability of a word and sensitive information jointly occurring. A sentence gets predicted as sensitive when this probability exceeds a given threshold. 

\subsubsection{(Bi-) Long Short Term Memory}\label{sec:LSTM}
In recent times, deep learning models have become the standard to detect and protect sensitive information. A specific deep learning model that is popular is called the Long Short Term Memory (LSTM) \cite{hochreiter1997long}.
Several sensitive information detection \cite{lin2020sensitive, wollmer2010bidirectional, neerbek2020real} works use the LSTM model \cite{hochreiter1997long} as it is capable of learning long-term dependencies. The LSTM model \cite{hochreiter1997long} is a special type of a RNN \cite{rumelhart1986learning}. Bi-LSTM models are even more sophisticated than the standard LSTM models. Bi-LSTMs train two LSTMs on the input sequence. The first input is the standard sequence and the other is a reversed copy of the first. This provides more context to the network and leads to more accurate results. To detect sensitive information the context is often crucial, however, the authors in \cite{neerbek2020real} show that LSTM gives no better results than the Inference Rule discussed in Section \ref{inference_rule}.

\subsubsection{Recursive Neural Network}\label{sec:recnn}
Another NLP model that has been tested to detect sensitive information is the Recursive Neural Network (RecNN). The RecNN \cite{rumelhart1986learning} is an ML model that is used for sequential data and is therefore suitable for text. RecNN has an internal memory as the output of each neuron is also used as the input. The RecNN has proved to be successful in paraphrasing detection in \cite{socher2011parsing} and therefore Neerbek \textit{et al.} tested it on the detection of sensitive information \cite{neerbek2020real}. The authors showed that RecNN gives no better results than the Inference Rule as discussed in Section \ref{inference_rule}.

\subsubsection{Transformers}\label{sec:transformers}
Transformers have become a popular architecture in the NLP field. Transformers are suitable for text and can handle long-range dependencies with ease. Vaswani \textit{et al.} \cite{vaswani2017attention} introduced pre-trained transformers in 2017. The biggest advantage of transformer models is that they can be parallelised, which reduces the training time significantly. In addition, transformers outperform the standard RNN and CNN models when tested on common NLP tasks such as GLUE, SQuAD and SWAG \cite{devlin2018bert}. Recently, transformers pre-trained on large corpora and fine-tuned for specific language tasks have shown to be effective. Therefore, a lot of NLP research has focused on transformers over the last few years.

Well-known language models based on transformers are BERT \cite{devlin2018bert}, Generalized Autoregressive Pretraining for Language Understanding (XLNet) \cite{yang2019xlnet}, Robustly optimized BERT approach (RoBERTa) \cite{liu2019roberta}, A Lite BERT (Albert) \cite{lan2019albert}, the Text-to-Text Transfer Transformer (T5) \cite{raffel2019exploring} and a distilled version of BERT (DistilBERT) \cite{sanh2019distilbert}. All these different NLP transformers have their own advantages. The original English-language BERT model is trained on the BooksCorpus consisting of 800M words and English Wikipedia consisting of 2,500M words \cite{devlin2018bert}. Albert has fewer parameters than BERT and the authors claim to have almost the same performance on the benchmark tasks. RoBERTa is trained on a bigger corpus than BERT and therefore, achieves better results on the benchmark tasks\footnote{RoBERTa is trained on the same 16GB corpus of BERT plus 144GB extra data.}. XLNet is also trained on a bigger corpus than BERT and performs better on the benchmark tasks\footnote{XLNet is trained on the same 16GB corpus of BERT plus 97GB extra data.}. DistilBERT has half of the parameters of BERT and is, therefore, faster but also less accurate than BERT.

Transformers have been applied successfully to detect sensitive information for unstructured data \cite{pablos2020sensitive, guo2021exsense}, but have not yet been tested on complex sensitive information such as in the Monsanto data set. In this work, we investigate whether transformers can also work for complex sensitive information. 

\section{Proposed Model} \label{proposed_model}

\begin{figure}[tbp]
\centerline{\includegraphics[width=.9\linewidth]{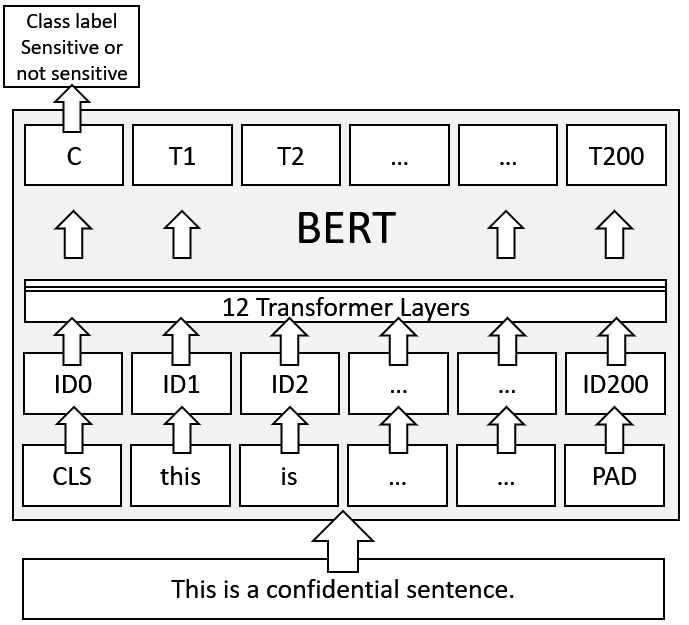}}
\caption{Visualisation of sensitive information detection with the BERT transformer.}
\label{fig:bert}
\end{figure}

We will detect sensitive information by fine-tuning a pre-trained BERT model \cite{devlin2018bert}. We use the original English language BERT model that is pre-trained on the BooksCorpus \cite{zhu2015aligning} and Wikipedia. This is the most well-known NLP transformer and has proven to be successful on different tasks such as paraphrase detection on the Microsoft Research Paraphrase Corpus \cite{devlin2018bert,dolan2005automatically}. As our task is similar to paraphrase detection, we will use the original English language BERT model for our task. The BooksCorpus contains 800M words and the English Wikipedia 2,500M words. From the Wikipedia corpus lists, headers and tables are ignored to ensure that long continuous sequences can be trained.

We pre-process the Monsanto data set by applying the bert-base-uncased tokeniser. This is the same tokeniser that was used during the pre-training of the BERT model on the BooksCorpus and Wikipedia. We use a maximum sequence length of 200 as none of the sentences in the Monsanto data set are longer than 200 words. Limiting the sequence length also reduces the training time. As we are working on a classification task it is crucial for the BERT tokenisation process to include the special [CLS] token as shown in Figure \ref{fig:bert}. [CLS] stands for classification and this special token gets inserted at the beginning of every sentence. As a last step of the tokenisation process, all words get converted to ID integers given by the tokeniser vocabulary which means that every sentence will start with the same ID integer representing the special [CLS] token. If any of the tokens do not appear in the vocabulary they will be assigned a special unknown token, [UNK], and the corresponding ID integer. Any sentence shorter than 200 words will be complemented with the [PAD] special token till the 200-word limit is reached.

We use the BertForSequenceClassification bert-base-uncased pre-trained model \footnote{Link: \url{https://huggingface.co/transformers/model_doc/bert.html\#bertforsequenceclassification}}. This model was trained using the masked language modelling (MLM) objective. During the pre-training phase, tokens from the input were randomly masked with the goal of predicting the masked tokens based on their context. The BertForSequenceClassification bert-base-uncased pre-trained model does not distinguish between lower case and upper case letters which is a common approach for English NLP tasks. The BERT base model consists of one encoder which consists of 12 layers of connected Transformers. We use the BERT base model instead of the BERT large model as the size of our data set is relatively small. 

As we are ultimately working on a sentence classification task, the sentence sequence will ultimately be reduced to a single vector. For the BERT classification task, only the hidden state of the first [CLS] gets selected at the end. {\color{black} Selecting the hidden state of the [CLS] token instead of max or average pooling is a critical component of the BERT method.} Ultimately, we classify the sentences as sensitive or non-sensitive with a linear classifier. 

It is observed in \cite{zhang2020revisiting} that the output of the BERT model, especially for small data sets, is highly dependent on the random seed chosen during the fine-tuning stage of a pre-trained BERT model. We will also analyse whether the fine-tuned pre-trained BERT model suffers from this type of instability. 

\section{Evaluation}\label{evaluation}
Section \ref{data_set} discusses the data set we use for our experiments, the metrics used for evaluations are presented in Section \ref{metrics}, Section \ref{results} summarises the results of our data analysis and Section \ref{discussion} higlights some interesting discussion points.

\subsection{Data Set}\label{data_set}

We evaluate the performance of our sensitive information detection model based on documents released as part of the Monsanto trial \cite{mchenry2018monsanto}. The Monsanto trial started in 2017 as a consequence of Monsanto being accused of claiming that their weed killer is safe while knowing that it can cause cancer. The documents released are internal Monsanto documents, in English, many of which contain confidential information about weed killer safety.

This data set is particularly well suited for sensitive information detection analysis. This is not only a real data set but is also labelled. It is hard to find real-world data sets of this kind as sensitive documents are often held secret due to their nature. Before the Monsanto data became available, the Enron corpus was the primary data set used for sensitive information detection research\cite{klimt2004enron, cormack2010overview}. Unlike the Monsanto data set, the Enron data set is unlabelled and therefore less suitable for this work.

The corpora is divided into four categories:
\begin{itemize}
 \item GHOST: Documents related to Monsanto employees ghostwriting and peer-reviewing scientific articles and attempts to influence journals to revoke studies harming Monsanto.
 \item TOXIC: Documents related to the toxicity of a Monsanto product.
 \item CHEMI: Documents on the chemistry of a Monsanto product.
 \item REGUL: Documents related to government and regulatory influence.
\end{itemize}

This Monsanto data set has {\it silver} and {\it golden} labels. The {\it silver} labels were provided by the lawyers at Baum, Hedlund, Aristei \& Goldman. The {\it golden} labels were provided by three annotators as part of the study of \cite{neerbek2020real} \footnote{Link to labelled data: http://lrec2020.lrec-conf.org/sharedlrs2020
/219\_res\_1.zip. trees0 refers to GHOST, trees1 to TOXIC, trees3 to CHEMI and trees4 to REGUL.}.

{\it Golden} labels are sentence-level: every single sentence gets a binary label indicating whether or not it contains sensitive information.\footnote{The term {\it golden} and {\it silver} labels come from \cite{neerbek2020real}.} {\it Golden} labels require experts to label each sentence in a corpus, so it is labour intensive, expensive and consequently difficult to obtain. 

We refer to  document-level sentence labels as the {\it silver} labels. All sentences in a document receive the label of the document. If a document contains any kind of sensitive information all sentences get labelled as sensitive. If a document does not contain any kind of sensitive information all sentences get labelled as non-sensitive. While the {\it silver} labels require less labour, the disadvantage is that this method is less discriminating. For example, a sentence that is non-sensitive, but appears in sensitive documents will be incorrectly labelled as sensitive. In this work, we work with the {\it golden} labels as these are more accurate.

\begin{table}[tbp]
\caption{Golden data set. The first row of each cell is the total number of labelled sentences and the second row is the number of sentences labelled as sensitive. The third row of each cell shows the percentage of sentences labelled as sensitive.}
\label{tab:data_golden_breakdown}
\begin{center}
\begin{tabular}{|c|c|c|c|c|}\hline
 & \textbf{Train} & \textbf{Validation}& \textbf{Test}& \textbf{Total} \\ \hline
 
 & 144 & 62 & 90 & 296\\ \cline{2-5} 
GHOST& 41& 14 & 22 & 77\\ \cline{2-5} 
& 28.47\% &22.58\%& 24.44\%& 26.01\% \\\hline

 & 134 & 65 & 53 & 252 \\ \cline{2-5} 
TOXIC& 26& 15 & 16 & 57\\ \cline{2-5} 
&19.40\% &23.08\% &30.19\%&22.62\%\\ \hline

 & 123& 61& 66& 250 \\\cline{2-5} 
CHEMI& 17& 5& 10 & 32\\\cline{2-5} 
&13.82\%& 8.20\% &15.15\%&12.80\% \\ \hline 

 & 139 & 69 & 67 & 275 \\\cline{2-5} 
REGUL& 19& 9& 6& 34\\ \cline{2-5} 
& 13.67\% &13.04\% &8.96\%&12.36\% \\\hline

& 540 & 257& 276& 1073\\\cline{2-5} 
Total & 103 & 43 & 54 & 200\\\cline{2-5} 
&19.07\%& 16.73\%& 19.57\%& 18.64\% \\\hline
\end{tabular}
\end{center}
\end{table}

\begin{table}[b!]
 \caption{The percentage split of the training, validation and test data of the Monsanto data set.}
 \label{tab:perc_breakdown}
 \begin{center}
 \begin{tabular}{|c|c|c|c|} \hline
 & \textbf{Train} & \textbf{Validation} & \textbf{Test} \\ \hline
 GHOST & 48.65\% &20.95\%&30.41\% \\ \hline
 TOXIC &53.17\%&25.79\%&21.03\%\\ \hline
 CHEMI &49.20\%&24.40\%&26.40\%\\ \hline
 REGUL &50.55\%&25.09\%&24.36\%\\ \hline
 \end{tabular}
 \end{center}
\end{table}

The annotators distinguish the documents into four categories. Table \ref{tab:data_golden_breakdown} 
shows the composition of the golden 
data set with the breakdown of GHOST, TOXIC, CHEMI and REGUL. Table \ref{tab:perc_breakdown} shows the percentage split of the training, validation and test sets. For our experiments, we use the same training, validation and test sets as \cite{neerbek2020real}. As identified in the table, around 50\% of the data will be used for training, approximately 25\% will be used for validation and the remaining for testing.

\begin{table*}[tbh!]
\caption{\label{tab:overview}
Results for BERT and existing sensitive information detection models based on the Monsanto data set. All the sub data sets, GHOST, TOXIC, CHEMI and REGUL are analysed separately. For some of the existing sensitive information detection models, we have not been able to retrieve or reproduce the values as in \cite{neerbek2020real}.
}
\begin{subtable}[t]{0.4\textwidth}
\centering
\begin{tabular}{|c|c|c|c|c|c|}\hline
 & \textbf{Accuracy} & \textbf{Precision} & \textbf{Recall} & \textbf{F1} & \textbf{F2} \\ \hline
Inference Rule&76.67 & 40.00& 14.00& 20.74 & 16.09 \\\hline
LSTM & 75.56 & 00.00 &00.00 &00.00&00.00 \\\hline
RecNN & 75.56 & 00.00 &00.00 &00.00&00.00 \\ \hline
BERT& 84.44&{100}&{36.36} &53.33&{41.66} \\ \hline
\end{tabular}
\caption{\label{tab:GHOST}GHOST}
\end{subtable}
\hspace{1.7cm}
\begin{subtable}[t]{0.4\textwidth}
\centering
\begin{tabular}{|c|c|c|c|c|c|}\hline
 & \textbf{Accuracy} & \textbf{Precision} & \textbf{Recall} & \textbf{F1} & \textbf{F2} \\ \hline
Inference Rule&73.58 & 62.50 & 31.25 &41.67 &34.72 \\\hline
LSTM &69.81 & 00.00 &00.00 &00.00 &00.00 \\\hline
RecNN &67.92 & 00.00 & 00.00 & 00.00 & 00.00 \\ \hline
BERT&{75.47} & {71.43} &{31.25} &{53.57}&{60.00}\\ \hline
\end{tabular}
\caption{\label{tab:TOXIC}TOXIC}
\end{subtable}
\vspace*{0.5cm}

\begin{subtable}[b]{0.4\textwidth}
\centering
\begin{tabular}{|c|c|c|c|c|c|}\hline
 & \textbf{Accuracy} & \textbf{Precision} & \textbf{Recall} & \textbf{F1} & \textbf{F2} \\ \hline
Inference Rule&84.85 & 00.00 & 00.00 &00.00 &00.00 \\\hline
LSTM &84.85 & 00.00 &00.00 &00.00&00.00 \\\hline
RecNN &84.85 & 00.00 &00.00 &00.00&00.00 \\ \hline
BERT&84.85 &00.00 &00.00 &00.00&00.00\\ \hline
\end{tabular}
\caption{\label{tab:CHEMI}CHEMI}
\end{subtable}
\hspace{1.7cm}
\begin{subtable}[b]{0.4\textwidth}
\centering
\begin{tabular}{|c|c|c|c|c|c|}\hline
 & \textbf{Accuracy} & \textbf{Precision} & \textbf{Recall} & \textbf{F1} & \textbf{F2} \\ \hline
Inference Rule&91.04 & 00.00 & 00.00 & 00.00 &00.00 \\\hline
LSTM &91.04 & 00.00 &00.00 &00.00&00.00 \\\hline
RecNN &86.57 & 00.00 & 00.00 & 00.00 & 00.00 \\ \hline
BERT&{94.03} & {75.00} &{50.00} &{60.00}&{53.57}\\ \hline
\end{tabular}
\caption{\label{tab:REGUL}REGUL}
\end{subtable}
\end{table*}

\subsection{Metrics} \label{metrics}

We compare the results of the different sensitive information detection models based on five different metrics: Accuracy, Precision, Recall, F1 and F2. Accuracy is simply the ratio of correct sensitivity predictions:

\begin{equation}
Accuracy = \frac{TP + TN}{TP + TN + FP + FN}
\end{equation}

with $TP$ being true positive, $TN$ true negative, $FP$ false positive and $FN$ false negative.

Accuracy is of limited use as a metric in this case. Our data sets are unbalanced as shown in Table \ref{tab:data_golden_breakdown}, because the number of sentences labelled as sensitive is small. If a model predicts all sensitives to be non-sensitive a high Accuracy can be achieved, but this is not desirable as we are mainly interested in detecting sensitive sentences.

The Precision and Recall scores are more important for the detection of sensitive information than the Accuracy. The Precision is the ratio of true positives and the sum of the true positives and false positives:

\begin{equation} \label{eq:Precision}
Precision = \frac{TP}{TP+FP}
\end{equation}

The Precision score is high when lots of sentences are correctly predicted as sensitive relative to the total number of sentences predicted as sensitive. The Recall score, also known as the sensitivity, is the ratio of correctly predicted positive observations to all the positive cases:

\begin{equation} \label{eq:Recall}
Recall = \frac{TP}{TP+FN}
\end{equation}

The Recall is high when the many sentences get correctly predicted as sensitive relative to the total number of actual sensitive sentences. For sensitive information, the Precision and the Recall metrics are both valuable. We can combine both metrics and create the $F_\beta$ score:
\begin{equation} \label{eq:fbeta}
F_\beta = \frac{(1+\beta) * Precision * Recall}{\beta^2* Precision + Recall}
\end{equation}

The $F_\beta$ score is the weighted average of Precision and Recall. When $\beta=1$ the weight on Precision and Recall is balanced. The lower $\beta$ the more weight is assigned on Precision and the less weight on Recall. The higher $\beta$ the more weight is assigned on Recall and the less weight on Precision. In this work, we analyse $F_\beta$ with $\beta=1$ and $\beta=2$.

\subsection{Results}\label{results}

In this section, we show the results of applying the BERT model on all the four subsets, GHOST, TOXIC, CHEMI and REGUL, of the Monsanto data set.

We fine-tune this BERT model during the training phase. As the training data is small, e.g., the minimum number of sentences is 123 for CHEMI and the maximum is only 144 for GHOST, we apply an early stopping epoch of 3, a dropout rate of 0.1 and a maximum number of epochs of 10 to avoid overfitting. We experimented with different learning rates, 1e-5, 2e-5, 3e-5, 4e-5 and 5e-5 as these are the recommended fine-tuning learning rates for BERT by \cite{turc2019}. As our data sets are relatively small we opt for small batch sizes, the number of training examples utilised in one iteration, of 4 or 8 We use AdamW as the optimiser. AdamW is a variation of Adam and reduces the training time and can better generalise compared to Adam \cite{loshchilov2017decoupled}. All our models are run on four 12GB memory Nvidia GE Force RTX 2080 GPUs \cite{nvidia2021gpu}. The run time of each model is between 30 and 120 seconds.

The fine-tuned BERT model successfully distinguishes sensitive and non-sensitive information. Table \ref{tab:overview} shows how the fine-tuned BERT model compares to existing sensitive information detection models. Table \ref{tab:overview} shows the Accuracy, Precision, Recall, F1 and F2 scores compared to existing sensitive information detection models which are Inference Rule, LSTM and RecNN models as in \cite{neerbek2020real}. 

For all the four sub data sets, our fine-tuned pre-trained BERT model has the same or a higher score for all the five different metrics. This suggests that it is not only more accurate in prediction but also accurately detects sensitive information. As opposed to the LSTM and the RecNN with zero true positives, the fine-tuned pre-trained BERT model manages to correctly identify sensitive sentences. The Inference Rules correctly identifies a couple of sentences, but the fine-tuned pre-trained BERT model identifies many more. While the Inference Rule reports a Precision of 40.00, 62.50, 0.00 and 75.00 respectively for the GHOST, TOXIC, CHEMI and REGUL data sets, the fine-tuned pre-trained BERT model reports a Precision of 100, 71.43, 0.00 and 75.00. 

As discussed in Section\ref{metrics} we mainly value the results of the F2 metric. For the GHOST data set, the fine-tuned pre-trained BERT model reports an F2 of 41.66 compared to 16.09, 0.00, 0.00 for the Inference Rule, LSTM and RecNN subsequently. Also, for the TOXIC and REGUL data sets the F2 for the pre-trained BERT model is significantly higher compared to the Inference Rules, LSTM and the RecNN model.

None of the models succeeds to detect sensitive sentences of the CHEMI data sets. A potential reason is that the CHEMI training and validation sets are the smallest of all the subsets with respectively 123 and 61 sentences as shown in Table \ref{tab:data_golden_breakdown}. Another potential reason is that the CHEMI data set is skewed with only 13.85\% and 8.20\% of the sentences labelled as sensitive for the train and validation set. 

\begin{figure*}[tbh!]
\centerline{\includegraphics[width=.9\linewidth]{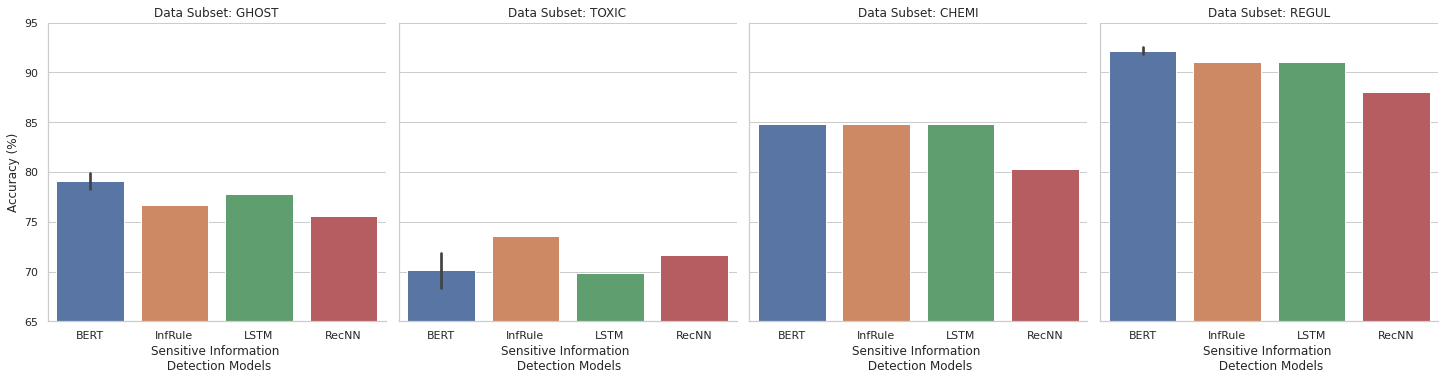}}
\caption{The accuracies for different sensitive information detection models. For the GHOST, TOXIC and REGUL sub-datasets the 95 percent confidence of the BERT model is shown.}
\label{fig:boxplots_golden}
\end{figure*}

We also analyse the robustness of the BERT model. Figure \ref{fig:boxplots_golden} shows the 95 percent confidence interval of the BERT accuracies when different seeds are used during the fine-tuning stage. The results of the BERT model is not very robust. The instability of the BERT model is a known issue and has been extensively analysed previously by \cite{zhang2020revisiting}. When fine-tuning a pre-trained BERT model we recommend fine-tuning with different seeds to evaluate the robustness of the model for the specific data set in use.

In our BERT model, the early stopping epoch, which means that the fine-tuning automatically stops when a chosen metric stops improving. The results in Table \ref{tab:overview} were based on an early stopping epoch depending on the Accuracy metric. We also experimented by letting the early stopping epoch depend on the F2 metric, so when the F2 worsens, the fine-tuning stops. Table \ref{tab:golden_label_results_details_f2} shows the results of the Accuracy, Precision, Recall, F1 and F2 score when the early stopping epoch of 3 is based on the F2 metric. The batch size for GHOST, TOXIC, CHEMI and REGUL are subsequently 4, 8, 8 and 8. The batch size is relatively small as the data set size is also relatively small. The learning rates are subsequently 5e-5, 3e-5, 3e-5 and 3e-5. The GHOST data set F2 increases from 41.66 to 65.79. The TOXIC data set F2 decreases from 60.00 to 54.88. The CHEMI data set F2 increases from 0.00 to 39.22. The REGUL data set F2 stays the same at 53.57. By letting the early stopping epoch depend on the F2 we can on average increase the F2 score. 

\begin{table}[t!]
\centering
\caption{\label{tab:golden_label_results_details_f2} Accuracy, Precision, Recall and F1 of golden data set when we train BERT based on F2}
\begin{tabular}{|c|c|c|c|c|c|}
\hline
& \textbf{Accuracy} & \textbf{Precision} & \textbf{Recall} & \textbf{F1} & \textbf{F2} \\ \hline
GHOST & 80.00&57.69&68.18&62.50&65.79\\ \hline
TOXIC & 69.81&50.00&56.25&52.94&54.88\\ \hline
CHEMI &80.30&36.36&40.00 &38.10 &39.22\\ \hline
REGUL &94.03&75.00 &50.00 &60.00 &53.57\\ \hline 
\end{tabular}
\end{table}

\subsection{Discussion} \label{discussion}

A BERT transformer was fine-tuned on the Monsanto data set. Our results showed that the BERT model returned higher Precision, Recall, F1 and F2 scores compared to existing sensitive information detection models. We observed that existing sensitive information detection models often ended up predicting all sentences as non-sensitive.

We also showed that the BERT models achieve on average a higher Precision and Recall score when the early stopping epoch during the fine-tuning of BERT is based on F2. This implementation has huge benefits for many practical applications. 

The results showed that the detection Accuracy of sensitive sentences is dependent on the random seed used during the fine-tuning stage. This is in line with what we expected based on the research \cite{zhang2020revisiting} where it was shown that a fine-tuned BERT model is unstable, especially when small data sets are used.

The main reason for the better performance of the BERT model compared to existing sensitive information detection models is that BERT is a pre-trained transformer. Our method benefits from the knowledge that BERT builds upon as it was pre-trained on Books Corpus and English Wikipedia. This means that the BERT model has already learnt about the structure, grammar and other linguistic features of texts. 

An advantage of using a pre-trained transformer like BERT, as opposed to some other ML models, is the speed. {\color{black}The training time is significantly lower as only fine-tuning is required compared to training a model from scratch.} This speed is further improved as transformers can be trained on multiple GPUs simultaneously which reduces the learning time significantly. 

We expect that in the future that more advanced transformers will be released which will increase the performance even further. Whilst writing this paper, Google announced that they developed a new transformer, named MUM \cite{google2021mum}. The authors claim that MUM is 1,000 times more powerful than BERT.

The Monsanto Trial data set is a realistic and high quality labelled data set. The set contains individually labelled sentences that contain complex information. Although the quality, realism and granularity of the data set is high, the data set size is small. The smallest golden data set is CHEMI, which only has a size of 250. The biggest, GHOST, is not much bigger at 296. The size of this data set limits the experimental opportunities. For example, we are unable to analyse the effect of the data set size on the performance of the different sensitive information detection models. We encourage the creation of a data set that has the same quality, granularity and realism level as the Monsanto trial data set but has a higher volume. Creating such a data set is challenging as sensitive information is generally kept secret due to its nature.

In short, this work showed that a fine-tuned BERT model outperforms existing sensitive information detection models on common metrics like Precision, Recall, F1 and F2. We also showed that by changing the evaluation metric during the fine-tuning process the Precision, Recall, F1 and F2 scores can be further improved. 

\section{Conclusion}\label{sec:conclusion}
In this work, we explored whether complex sensitive information can be detected by fine-tuning pre-trained transformers. This was tested by fine-tuning a pre-trained BERT transformer model and comparing its performance when test on the GHOST, TOXIC, CHEMI and REGUL data sets of the Monsanto Trial with the Inference Rule, RecNN and LSTM models.

Previous analysis of the performance did not investigate the Precision, Recall, F1 and F2 scores, which we argue is more pertinent to ensure the detection of sensitive information is accurate. Specifically, we believe that the F2 score is the most important as this is a weighted average of the Recall and Precision with a higher weight for the Recall. The Recall is important as it measures the ratio of sentences correctly predicted as sensitive and the total number of actual sensitive sentences. 

In all instances, BERT returned a higher F2 score. This implies that fine-tuning pre-trained transformer models will perform better than existing sensitive information detection models. In conclusion, for sensitive information detection in the industry, we recommend the use of fine-tuned pre-trained transformer models. 

\section{Acknowledgements}
The authors acknowledge the support of the Commonwealth of Australia, Cyber Security Research Centre Limited, and Penten for this work.

\printbibliography

\end{document}